\documentclass[10pt,twocolumn,letterpaper]{article}

\usepackage{cvpr}
\usepackage{times}
\usepackage{epsfig}
\usepackage{graphicx}
\usepackage{amsmath}
\usepackage{amssymb}

\usepackage[breaklinks=true,bookmarks=false]{hyperref}

\cvprfinalcopy 

\def\cvprPaperID{****} 

\begin{document}

\title{DriftNet: Aggressive Driving Behavior Classification using 3D EfficientNet Architecture}

\author{Alam Noor$^{1}$, Bilel Benjdira$^{1}$, Adel Ammar$^{1}$, Anis Koubaa$^{1,2}$\\
	{$^{1}$Robotics and Internet-of-Things Lab (RIOTU), Prince Sultan University, Riyadh, Saudi Arabia.} \\
	{$^{2}$CISTER, INESC-TEC, ISEP, Polytechnic Institute of Porto, Portugal.}\\   
	nalam@psu.edu.sa, bbenjdira@psu.edu.sa, aammar@psu.edu.sa, akoubaa@psu.edu.sa
}

\maketitle

\begin{abstract}
Aggressive driving (i.e., car drifting) is a dangerous behavior that puts human safety and life into a significant risk. This behavior is considered as an anomaly concerning the regular traffic in public transportation roads. Recent techniques in deep learning proposed new approaches for anomaly detection in different contexts such as pedestrian monitoring, street fighting, and threat detection. In this paper, we propose a new anomaly detection framework applied to the detection of aggressive driving behavior. Our contribution consists in the development of a 3D neural network architecture, based on the state-of-the-art EfficientNet 2D image classifier, for the aggressive driving detection in videos. We propose an EfficientNet3D CNN feature extractor for video analysis, and we compare it with existing feature extractors. We also created a dataset of car drifting in Saudi Arabian context\footnote{https://www.youtube.com/watch?v=vLzgye1-d1k}. To the best of our knowledge, this is the first work that addresses the problem of aggressive driving behavior using deep learning.
\end{abstract}
\section{INTRODUCTION}

The problem of aggressive driving (also known as car drifting) represents a significant safety problem in several cities. It is often related to youngsters driving cars at high speed and drifting between other vehicles in the road, which put the safety and lives of other drivers into a significant threat. In this paper, we propose to use artificial intelligence for automated anomaly detection of abnormal driving behavior, specifically drifting. 

\paragraph{\textit{Related works}}
A variety of attempts were made in previous works to identify suspicious cars in videos automatically. Approaches based on video anomaly have mainly two feature extraction types (manual and deep learning). Various techniques have historically been carried out in statistical methods for the identification of anomalies. Irregular events, motion patterns, space, time, and texture of optical flow were detected using Hidden Markov Models \cite{1}, Gaussian Mixture Models, and Markov Random Fields \cite{2,3}. Various classification algorithms, such as Support Vector Machine\cite{4}, Nearest neighbor Clustering \cite{5}, and K-means clustering \cite{6} are used as a binary classification for anomaly detection to extract features and optical flow of regular and abnormal patterns of vehicles.

On the other hand, the approaches to deep learning differ from conventional methods. In the areas of object detection, the rapid implementation and growth of deep learning are primarily performed using a convolution neural network. CNN accomplished its achievement in detection by applying the divisional rule in one-stage to different small anchor boxes (grid) of various sizes \cite{7} and by using region proposal networks for the two-stage approaches \cite{8}, then they apply classification to labels and regression to bounding boxes. Different approaches of deep learning have been adopted at the bases of one-stage and two-stage methods to solve the anomaly detection problems. Wang et al. proposed the novel approach of Gaussian mixture models with the region of interest of the YOLOv3 to identify anomaly candidate and remove noises of the background. In last TrackletNet Tracker has been employed to obtain anomaly trajectory \cite{9}. Ammar et al. proposed the convolutional neural network based car detection method using Aerial Images with comparison of F-RCNN and YOLOv3 \cite{17}.  Shuai et al. presented the perspective method with the backend of ResNet50 and unsupervised traffic flow analysis to eliminate the moving car and detect the state objects on the road and the space-time matrix used to achieve anomaly car locations. \cite{10}. Belil et al., presented the Generative Adversarial Networks (GAN) architecture to reduce the domain shift with increasing the ability to new targeted domains \cite{18}, and using Faster R-CNN and YOLOv3 to detect the cars from drones with high precision \cite{19}. 
Jianfei et al. proposed an unsupervised framework with multi-object tracking to combine single-object tracking to solve the tracks and detection problems with efficiency and accuracy \cite{11}. Ding et al. employed 3D ConvNet's end-to-end architecture to detect violence in videos \cite{12}. Ji et al. presented 3D CNN using bottleneck technique with adopting the DenseNet method of several dense layers and in the last global average pooling layer for feature mapping \cite{13}. Google researchers Mingxing et al. proposed a new and scalable architecture EfficientDet to use a weighted bi-directional feature pyramid network with a combination of EfficientNet as a backbone following the one-stage detector design. Additionally, adopting optimize multi-scale feature fusion for interpretation of resulting feature network \cite{14}.   

In this paper, we leverage the use of the EffientNet scaling approach, and we adopt it as a backbone multimodal architectural framework for car drifting detection.

The rest of the paper is organized as follows. In Section 2, we present the DrifNet architecture, where we presented the dataset collection process and the proposed 3D EfficientNet neural network model.
\section{DriftNet Architecture}
In this section, we provide the details of the pipeline of the proposed method for driving anomaly detection. Anomalies of traffic in the AI City Challenge issue are primarily vehicles that have different behaviors from other cars like drifting\footnote{https://www.youtube.com/watch?v=vLzgye1-d1k}, stopping, crashing on the road, etc. First, we created a dataset (described below) for vehicle drifting. Second, we proposed a 3D EfficientNet method to detect and recognize aggressive car driving efficiently. Using 3D EfficientNet to detect car drifting, it should be able to identify the pattern of moving vehicles in sliding modes and motions.  We considered EfficientNet because it is currently the state-of-the-art algorithm for object detection. Efficiency and accuracy are an essential part of the recognition tasks, and the models of EfficientNet achieve higher accuracy and performance over an extensive range of precision and resource constraints than previous detectors.     
\subsection{Drifting Dataset}
The compilation and annotation of large-scale datasets are painstaking, expensive and time consuming. We created a car drifting clips dataset to advance computer vision and video understanding of aggressive driving detection. 100 video samples of drifting in Saudi Arabia roads were collected from YouTube, and added to 100 videos of normal traffic selected from the AI Challenge dataset. We performed weakly labeling of the video frames as drifting and normal driving, and vocabulary annotation has been designed carefully. We extracted a total of 4900 frames for normal driving and 4900 frames for vehicle drifting.
We split the dataset into 80\% for training, and 20\% for validation. Table.1 shows the number and percentage of frames of the training and testing sets. Figure 1 shows sample frames of normal driving from the AI City Challenge dataset.
\begin{figure}[ht]
    \begin{center}
        \includegraphics[scale=0.4]{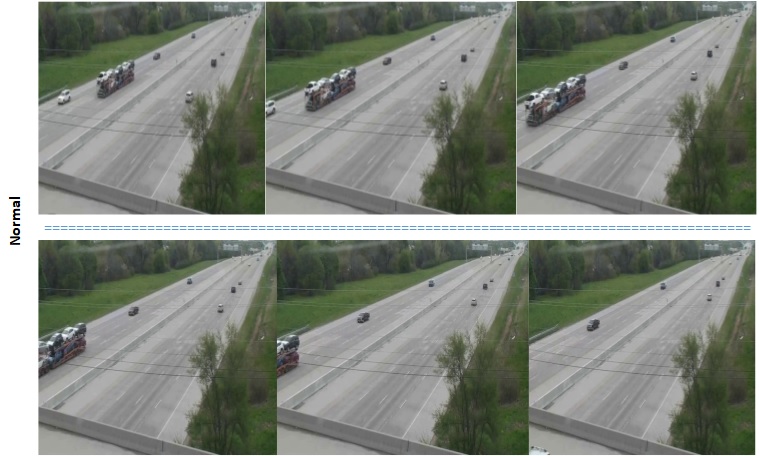}
        \caption{From top left to bottom right, all the cars shown are in normal driving positions.}
        \label{fig.1}
    \end{center}
\end{figure}
\begin{table}
\caption{Anomaly car driving detection.}
\label{TABLE I}
\centering
\hspace*{-3em}
\begin{tabular}{ |l| c |c |c|  p{1cm}}  
 \hline
  Categories & Training set  & Testing set& Total \\  
 \hline
 Number of images  & 7840& 1960 &9800  \\ 
 Percentage & 80\% &20\% &100\%   \\
 Number of drift &3920 &980	&4900	\\
 cars frames&&&\\
 Number of non-drift &3920	&980	&4900	\\
 cars frames&&&\\
 \hline
 \end{tabular}
\end{table}
Each frame has a size of 640x360. We pre-processed all frames to make them suitable for input size 112x112 for the EfficientNet feature extractor. This dataset will be made open-source to drive new research on car drifting detection so that it can effectively contribute to AI City challenges to increase human safety and mitigate the risks of aggressive driving behavior. Figure 2 shows sample frames of the aggressive driving car from the anomaly dataset.
\begin{figure}[ht]
    \begin{center}
        \includegraphics[scale=0.4]{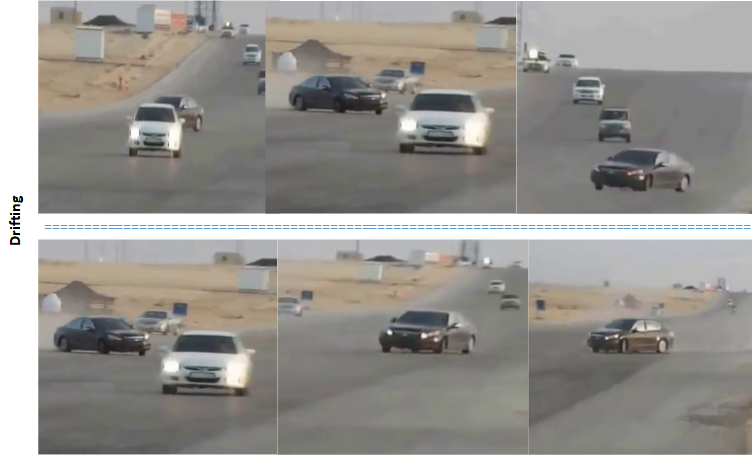}
        \caption{From top left to bottom right the black car shows different positions of drifting.}
        \label{fig.2}
    \end{center}
\end{figure}
\subsection{3D EfficientNet Network}
We evaluated EfficientNet using transfer learning for the drifting dataset. We borrow the same training settings from EfficientNet, which takes the ImageNet pre-trained model and finetune it to the new dataset. 
Figure 3 represents the workflow of the EfficientNet architecture.
\begin{figure*}[]
    \begin{center}
        \includegraphics[width=18cm]{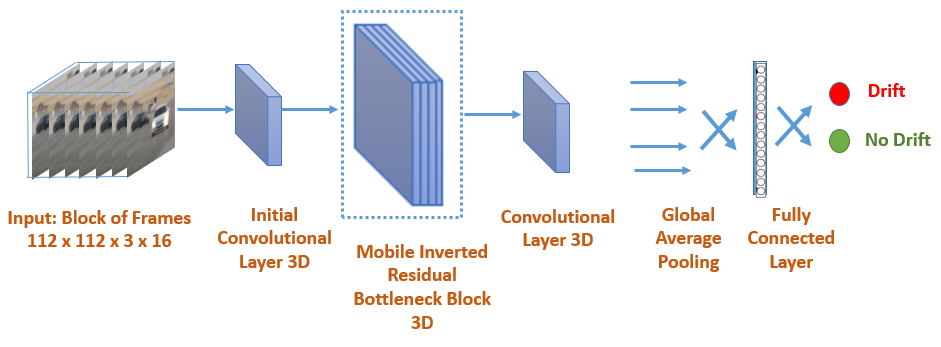}
        \caption{EfficientNet3D-B0 Architecture.}
        \label{fig.3}
    \end{center}
\end{figure*}
The model optimization algorithm 'SGD' was used during the training process with the following parameters: the starting learning ratio is 0.0002, the momentum is 0.5, and the decay rates are 1e-7. After every convolution neural network, Batch Normalisation is applied with a 0.997 decay batch norm. Besides, a batch of samples was 32 for batch testing and model validation. The sample size was 112 x 112. Besides, the early stop training technique was used to avoid overfitting of the model during the training cycle.
\section{Experimental results}
This section details the experimental results of the proposed method deep transfer learning method. The experiment was conducted on Nvidia GeForce RTX 2070 GPUs and used with PyTorch. We have used four different architecture for effectiveness and efficiency of EfficientNet at drift dataset.

The preparation of input shape dimensions are N x C x T x H x W. Where N and C are the batch size and Channels respectively. H and W are the height and width of the frames, while T is the duration of video of the dataset. Frames of each video has been sampled into 32 and resized to an input size of 112 x 112. It is necessary to have a large dataset to minimize the over fitting problems during training, and data augmentation is a very effective method to avoid over fitting. Horizontal flipping, lighting, cropping, slicing and salt and pepper noise techniques have been used for data augmentation. We used the Stochastic Gradient Descent (SGD) optimizer for training with momentum 0.5 and learning rate decay 1e-7. We split the drifting dataset into 80\% for training and 20\% for testing. Figure \ref{fig-accuracy} and \ref{fig-accuracy} depict the training and validation loss and accuracy of EfficientNet3D-B0. We have tested other scaling factors for EfficientNet (up to B7) but they did not produce better results.

\begin{table*}
\caption{Experiment results on car drifting dataset.}
\label{Table 2}
\centering
\hspace*{-4em}
\begin{tabular}{ l c c c c c p{5cm}}  
 \hline
 Models & Param & Training Accuracy (\%)& Validation Accuracy (\%)  \\  
 \hline
 EfficientNet3D-B0 &4.69M & 93.75 & 92.5\\
 \hline
 \end{tabular}
\end{table*}
We compared our proposed model to three other existind models, namely, ConvLSTM, DenseNet, and C3D. C3D \cite{15} is a traditional 3D CNN video descriptor, while ConvLSTM \cite{16} and DenseNet are state-of-the-art models for the identification of anomalies. The C3D model's parameters are initialized in the initial convolution layer, while ConvLSTM uses pre-trained AlexNet for extracting features, and DenseNet takes sequentially multiple layers with batch normalization to extract the features of the object. We have taken pre-trained and applied transfer learning because the scale of driving car dataset is still limited to allow for a complete training of all the parameters of these deep neural networks. 
The efficiency and parameter number of these models is compared in Table 2. The proposed model shows a higher accuracy in classification with comparatively less parameters. Our model's validation accuracy is 92.5\% 

\begin{figure}
    \begin{center}
        \includegraphics[width=8cm]{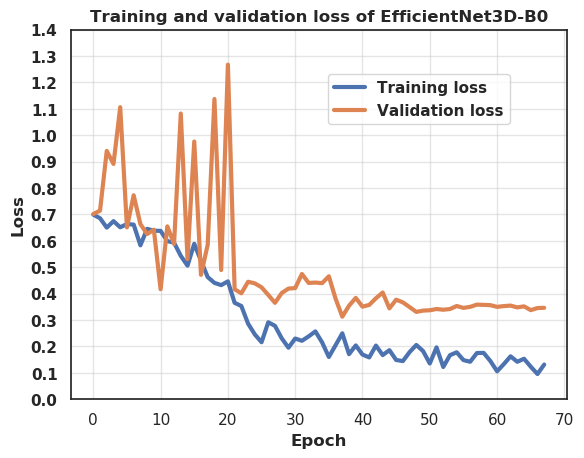}
        \caption{Training and validation loss of EfficientNet3D-B0.}
        \label{fig-loss}
    \end{center}
\end{figure}

\begin{figure}
    \begin{center}
        \includegraphics[width=8cm]{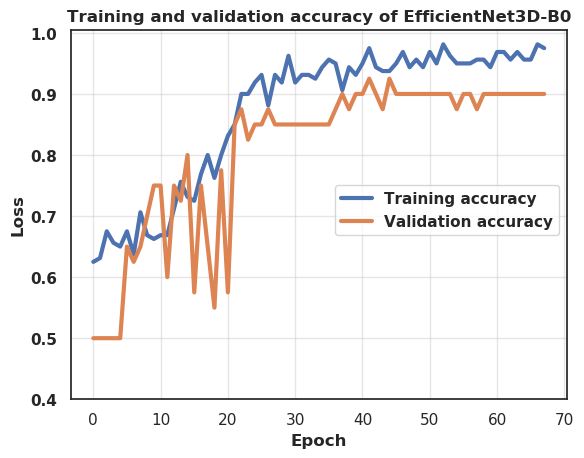}
        \caption{Training and validation accuracy of EfficientNet3D-B0.}
        \label{fig-accuracy}
    \end{center}
\end{figure}

\begin{figure}
    \begin{center}
        \includegraphics[width=8cm]{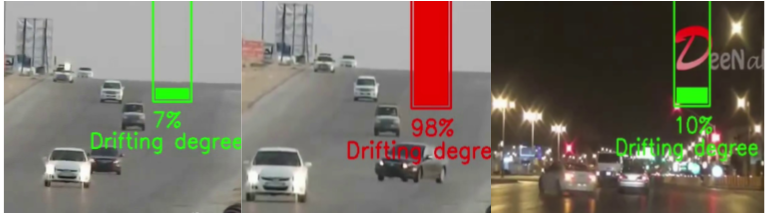}
        \caption{Output of the proposed architecture.}
        \label{Output}
    \end{center}
\end{figure}

\section{Conclusion}
This paper proposes an efficient network architecture design to improve the internal architecture as a transfer learning to efficiently detect anomaly vehicles. In order to increase accuracy and efficiency of the anomaly car detecting, the proposed model has comparatively lower parameters with a custom compound scaling process. The experimental results of car drift data sets are based on these optimizations and demonstrate the improvements of our model over framework. The proposed approach is to computing resource-saving and to achieve consistently greater precision and efficiency over a wide variety of limitations of resources than the existing state-of-the-art detection methods.

{\small
\bibliographystyle{unsrt}
\bibliography{biblio}
}

\end{document}